\pdfoutput=1

\documentclass[11pt]{article}

\usepackage[]{acl}

\usepackage{times}
\usepackage{latexsym}
\usepackage{float}
\usepackage{amsmath}
\usepackage{amsfonts}
\usepackage{amssymb}
\usepackage{graphicx}
\usepackage{caption}
\usepackage{subcaption}
\usepackage{color}
\usepackage{booktabs}
\usepackage{adjustbox}
\usepackage{makecell}
\DeclareUnicodeCharacter{0307}{}

\newcommand{\fae}{FILM}
\newcommand{\cnndm}{CNNDM$_{abs}$~}

\usepackage[T1]{fontenc}

\usepackage[utf8]{inputenc}

\usepackage{microtype}

\title{Faithful to the Document or to the World? Mitigating Hallucinations via Entity-linked Knowledge in Abstractive Summarization}

\author{Yue Dong$^1$\thanks{*This work was done when the first author was
an intern at Google Research.} \quad John Wieting$^{2}$ \quad Pat Verga$^2$ \\ \\
    $^1$Mila / McGill University \quad $^2$ Google Research\\
    {\tt yue.dong2@mail.mcgill.ca} \\
     \{\tt jwieting, patverga\}@google.com
	}

\begin{document}
\maketitle
\begin{abstract}
	Despite recent advances in abstractive summarization, 
	current summarization systems still suffer from content hallucinations where models generate text that is either irrelevant or contradictory to the source document. However, prior work has been predicated on the assumption that any generated facts not appearing explicitly in the source are undesired hallucinations. Methods have been proposed to address this scenario by ultimately improving `faithfulness' to the source document, but in reality, there is a large portion of entities in the gold reference targets that are not directly in the source. In this work, we show that these entities are not aberrations, but they instead require utilizing external world knowledge to infer reasoning paths from entities in the source. We show that by utilizing an external knowledge base, we can improve the faithfulness of summaries without simply making them more extractive, and additionally, we show that external knowledge bases linked from the source can benefit the factuality of generated summaries. 
\end{abstract}
\section{Introduction}

\begin{table}[t]
\small
\begin{tabular}{p{1.5cm}|p{5.3cm}}
\toprule
Source   & A fire crew remains at {\color{red} Plasgran} in Manea Road, {\color[HTML]{008000}Wimblington}, more than 16 hours after the incident began on Wednesday afternoon. Road closures are expected to stay in place until midday, the fire service said. About 75 firefighters worked into the night to put out the fire. They also prevented its spread to neighbouring properties. The incident was scaled down at 2300 GMT, when the fire was brought under control.  \\ \midrule

Summ. System Generation & A large fire has broken out at a {\color[HTML]{FF7E00}recycling centre} in {\color[HTML]{9932CC}Oxfordshire}, the fire service has said, forcing the closure of a road.  \\                             \midrule                
After Correction & A large fire has broken out at a {\color[HTML]{FF7E00}plastic recycling centre} in {\color[HTML]{007FFF}Cambridgeshire}...  \\                             \midrule   
Target     & An investigation has begun into the cause of a fire which has severely damaged a {\color[HTML]{FF7E00}plastics factory} in {\color[HTML]{007FFF}Cambridgeshire}. \\ \midrule
Relevant Facts Linked & 
1. {\color{red} Plasgran} $\rightarrow$ industry $\rightarrow$ {\color[HTML]{FF7E00}plastic recycling};  
2. {\color[HTML]{008000}Wimblington} $\rightarrow$ historic county $\rightarrow$ {\color[HTML]{007FFF}Cambridgeshire}; 
3. {\color[HTML]{008000}Wimblington} $\rightarrow$ also known as $\rightarrow$ {\color[HTML]{007FFF}Wimblington, Cambridgeshire}
\\
\bottomrule  
\end{tabular} 
\caption{The target summary contains out-of-article entities – {\color[HTML]{FF7E00}plastics factory} and {\color[HTML]{007FFF}Cambridgeshire} – that are important for comprehension. We can see that our model was able to correct the entities in the system-generated summary successfully with additional world knowledge linked from source entities (relevant facts). 
}
\label{tab:open_example}
\end{table}

\begin{figure*}[ht]
\centering
\includegraphics[width=.95\linewidth]{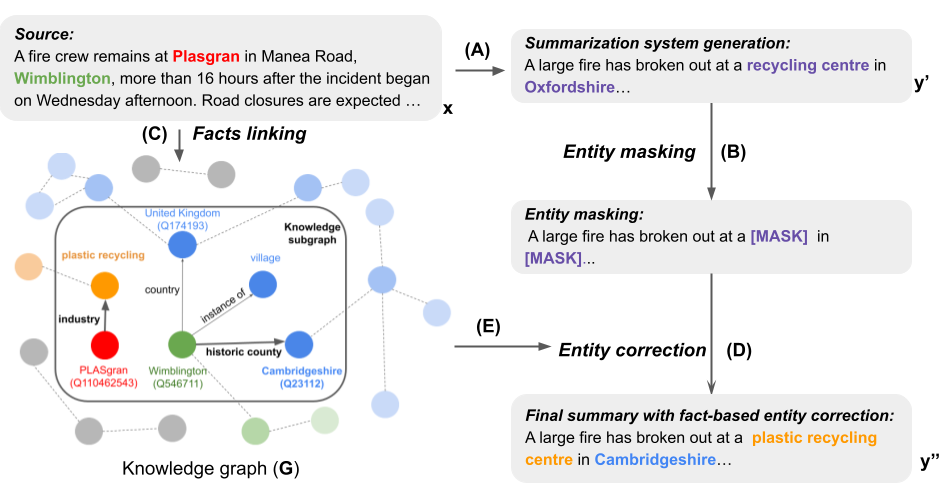}
\captionof{figure}{Schematic view of building the summarization pipeline with a knowledge enhanced entity correction. A) A standard seq-to-seq T5 model produces a generated summary. B) An entity linker is used to identify and mask out entities in the generated summary to produce a skeleton summary. C, D, E) The revision model (\fae{}) uses the source text, skeleton, and external knowledge base to revise and correct the masked entities.} 
\label{fig:overview1}
\end{figure*} 

Despite generating fluent summaries with high automatic evaluation scores, such as ROUGE \citep{lin-2004-rouge}, current state-of-the-art summarization models often hallucinate content that is not supported by the source \citep{pagnoni-etal-2021-understanding,maynez-etal-2020-faithfulness}. For instance, \citet{pagnoni-etal-2021-understanding} found that 92\% of generated summaries on XSUM \citep{narayan-etal-2018-dont} contain at least one factual error across different models tested. Even the best model – BertS2S – produces summaries with factual errors 83\% of the time.  In addition, based on their annotations, most of these factual errors are entity-based where generated summaries contain entities that are not in the source article (\textit{out of article entity error}). 

As far as we know, almost all prior works consider these \textit{out-of-article entities} generated by models as factually incorrect  \citep{falke-etal-2019-ranking,zhu-etal-2021-enhancing,pagnoni-etal-2021-understanding,nan-etal-2021-entity}, as they are not directly \textit{factual consistent} to the source. However, a large portion of these hallucinations, despite not being directly supported by the source, are factually correct; they are supported by additional world knowledge and often connect entities in the source to those used in the target. Moreover, some of these facts can provide additional information that is important to comprehend the summary \citep{ladhak2021faithful}. Table \ref{tab:open_example} shows an example  of this, where our method generates:  `Cambridgeshire' aligns to a target entity, and provides a clear explanation of where `Wimblington' is for readers who are not familiar with the
geography of UK.
In fact, our experiment on entity coverage (Figure~\ref{fig:entity_linking}) shows that 60\% to 71\% of entities, depending on the entity category in XSUM targets, are out-of-article entities that can not be found in the source.\footnote{Entity equivalence is resolved by linking surface forms to their Wikidata IDs \cite{vrandevcic2014wikidata}. For example, the string `US' and `United States' both map to the same Wikidata ID (Q30).} However, among different entity categories, 10.6\% to 31.6\% of these out-of-article entities can be found in the subgraph produced from a single hop beginning from the set of source entities mapped to the Wikidata knowledge base (KB) \cite{vrandevcic2014wikidata}. 

Since a large portion of target entities are exclusive to the external KB subgraph, one natural question to ask is: \textbf{can we leverage the external knowledge base to improve the faithfulness of summaries?} In this work, we focus on entities (e.g., person, event, location, organization) in summaries, as they are the most hallucinated \citep{pagnoni-etal-2021-understanding,kryscinski-etal-2020-evaluating} and often contain the most salient information. We treat the entities that are not in the source but in the external knowledge base linked from the source as faithful to the world knowledge.  Thus, we consider the faithfulness of a summary with respect to two aspects: faithful to the source document (extractiveness) and faithful to the world knowledge (abstractiveness).  Contrary to previous work that improves factual consistency by filtering training examples to contain only extractive entities \citep{nan-etal-2021-entity,narayan2021planning,mao2020constrained}, we focus on improving the faithfulness of the generated entities from the abstractive perspective: by providing additional facts that are relevant to the source.

Our contributions are:
\begin{itemize}
    \item We provide a comprehensive study on external knowledge bases connected from source entities in XSUM and \cnndm. For example, 59.9\% of location entities in the gold reference summaries of XSUM are not in the source;  31.6\% of these out-of-article entities can be found in the KB linked from the source document (Section \ref{sec:motivation}). 
    
    \item We explore multiple methods to incorporate the facts linked from the source to improve the faithfulness of the summaries, including a proposed revision model that revises entities in the generation to make them more consistent with the source consistent with the world knowledge (Section \ref{sec:methods}). 
    
    \item We propose entity-based metrics that evaluate the faithfulness of generated entities by matching their entity IDs from a knowledge base to those in the gold reference targets. This allows us to account for \textit{out-of-article entities} in our evaluation (Section \ref{sec:result}). 
\end{itemize}
\section{Case study: faithful to the document or the world? }
\label{sec:motivation}

The motivation of this work is the observation that many gold reference summaries in widely used summarization datasets, such as XSUM \citep{narayan-etal-2018-dont}, contain entities that do not appear explicitly in the source and instead require additional knowledge to resolve. We show that \textbf{much of this knowledge can be found by identifying facts in KBs that involve source entities}.

For a given example, we seek to supplement the source document with additional knowledge by constructing a subgraph derived from the Wikidata KB. We use Google Cloud NLU\footnote{\url{https://cloud.google.com/natural-language}} to identify typed entities in the source document as well as link those entities to their canonical form. This is necessary as the surface forms of entities often vary; for example, Wikidata ID Q30 - United States of America - has 18 different surface forms.\footnote{the United States of America | America | U.S.A. | USA | U.S. | US | the US | the USA | US of A | the United States | U. S. A. | U. S. | the States | the U.S. | 'Merica | U.S | United States | 'Murica} For each of these extracted entities, we collect the set of one-hop facts in Wikidata that originate at those entities. The set of all one-hop facts that begin at an entity appearing in our source document makes up our knowledge subgraph.

\begin{figure}[t]
	\centering
	\includegraphics[width=.99\linewidth]{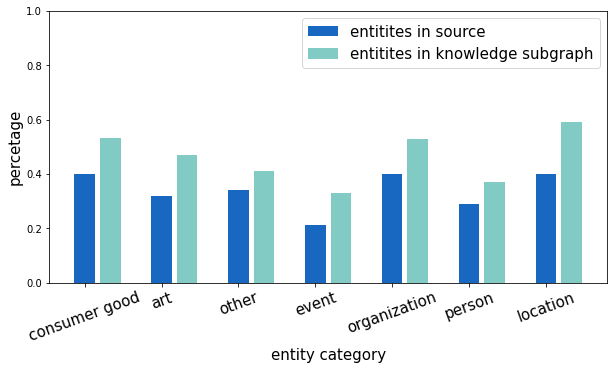}
	\caption{Increase of entity coverage by including external knowledge subgraphs. The knowledge subgraphs are constructed by including Wikidata facts that are one hop away from the set of entities in the source document.} 
	\label{fig:entity_linking}
\end{figure}

\begin{figure}[t]
	\centering
	\includegraphics[width=.99\linewidth]{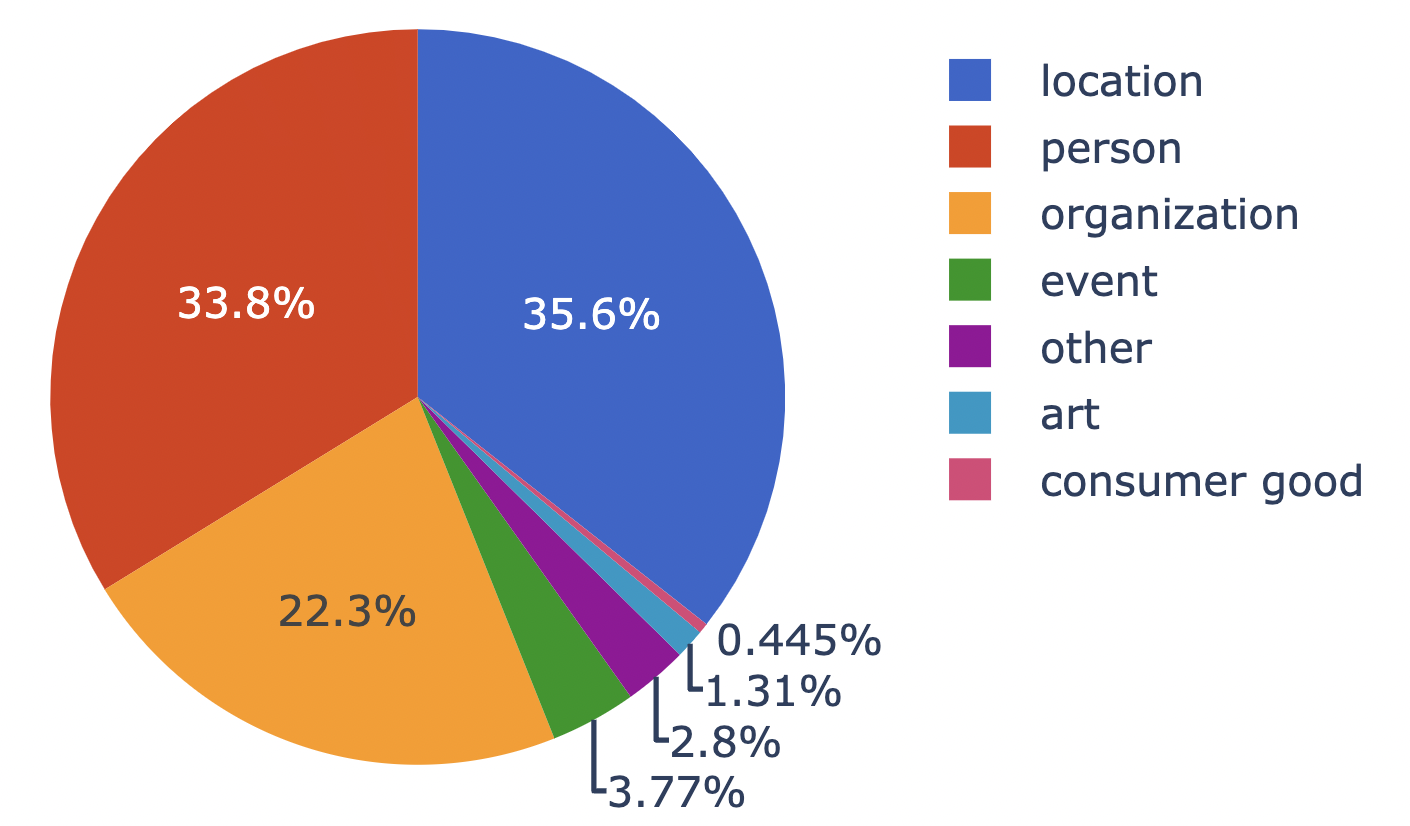}
	\caption{Entity proportion by type in XSUM.}
	\label{fig:entity_pie}
\end{figure}

Figure \ref{fig:entity_linking} shows that depending on the entity category,  60\% $\sim$ 79\% of target entities can not be found in the source in XSUM. However, a large portion of these so-called out-of-article entities can be found in the knowledge subgraph we constructed. Depending on the entity category,  8\% $\sim$ 19\% of target entities can be found \textit{exclusively} in the knowledge subgraph constructed from the set of source entities, resulting in 20.6\% $\sim$ 57.1\% improvement of entity coverage when compared to the set of source entities. 

Following single-hop KB links will not yield all related entities due to both the limited schema of the KB and the fact that the KB itself is highly incomplete \citep{lin-etal-2018-multi,ebisu-ichise-2019-graph}. However, relevant information can often be encoded in multi-hop paths through the graph ($Barack \ Obama \rightarrow born \ in \rightarrow Honalulu \rightarrow located \ in \rightarrow Hawaii$). Therefore, we also check whether an increase in the number of hops in linking would result in a higher coverage. In addition to XSUM, we extract an abstractive subset of CNN/Daily Mail dataset \citep{Hermann2015cnndm,nallapati-etal-2016-abstractive}, noted as CNNDM$_{abs}$, where at least one location entity in the target is not in the source. We obtained 95387/4357/3769 data instances for train/dev/test on \cnndm. 

\begin{table}[h]
	\resizebox{0.5\textwidth}{!}{%
		\begin{tabular}{lllll}
			\toprule
			Location      & \multicolumn{1}{|p{1.5cm}|}{Source Only} & \multicolumn{1}{p{1.5cm}|}{1 Hop}  & \multicolumn{1}{p{1.5cm}|}{2 Hops } & 3 Hops \\ \midrule
			XSUM          & \multicolumn{1}{|p{1.5cm}|}{40.1\%}      & \multicolumn{1}{p{1.5cm}|}{59.8\%} & \multicolumn{1}{p{1.5cm}|}{60.2\%}  & 60.3\% \\ 
			CNNDM$_{abs}$ & \multicolumn{1}{|p{1.5cm}|}{52.3 \%}     & \multicolumn{1}{p{1.5cm}|}{65.4\%} & \multicolumn{1}{p{1.5cm}|}{66.1\%}  & 66.2\% \\ \hline
		\end{tabular}
	}
	\caption{Target entity coverage after including facts from different number of hops beginning from source entities of the KB.}
	\label{tab:entity_linking}
\end{table}

Table~\ref{tab:entity_linking} shows most of the facts that are needed to connect source entities and abstractive target entities are within one hop of the KB. The benefits of traversing the KB with more hops to create the knowledge subgraph would result in negligible entity coverage gain (Table~\ref{tab:entity_linking}), and significantly more facts. For example, the average number of facts in the knowledge subgraph in XSUM after one hop of traversing is 790, and this number drastically increases to 5365 after two hops of traversing. 

In summary, contrary to the common beliefs that a summary is meant to be factually consistent with the source text, and therefore the source contains all the information in the summary; we have shown in this section that a large portion of gold references in XSUM and \cnndm require external knowledge. We provide one way to discover this external knowledge by linking the facts from the set of source entities and have shown that this knowledge subgraph provides a reasonable increase of target entity coverage. 

However, we do not claim that this is the best approach to construct the external knowledge subgraph for summarization. Indeed, we can see that despite the additional facts provides a boost in coverage of 19\% in location entities, there is still a large portion of target entities that are neither in the source nor in the knowledge subgraph. 
This can be due to multiple reasons. First, like all KBs, our seed KG (Wikidata) is not complete -- as entries are provided by users from manual edits, many links are missing \citep{min-etal-2013-distant}. In addition, the facts required to summarize XSUM and \cnndm can be temporally misaligned to the KG. For example in Wikidata, the President of the United States   (Q11696) is Joe Biden (2021 - present); but this entity could refer to different people depending on the years of the news article. Additionally, different KBs and subgraph selection methods could increase the entity recall further while reducing the number of spurious links, however, we leave this exploration for future work.

Nevertheless, our study indicates that humans would consider commonsense or external knowledge into their summary writing. In the rest of the paper, we explore whether incorporating explicit external knowledge would help reduce content hallucination from summarization models.  

\section{Methods}
\label{sec:methods}
In this section, we explore how to incorporate the additional knowledge into the model to reduce content hallucination and produce more faithful summaries. We introduce two methods to incorporate external knowledge in summarization: $(i)$ directly concatenate the knowledge subgraph to the source; $(ii)$ introduce relevant facts as an entity revision model.  In all of the experiments, we use T5-3B \cite{raffel2020exploring} as the base summarization model, the models are trained with each setting for 200k steps with a bath size of 128 on Cloud TPU v3 Pod with 128-core Pod slice. We use a constant learning rate of 1e-4 when fine-tuning. We save
a checkpoint every 5,000 steps and report results on the model checkpoint corresponding
to the highest validation performance. For entity extraction and linking with the types, we use Google Cloud NLU API. For fact-based entity correction, we finetune a fact-aware model - FILM \citep{verga-etal-2021-adaptable} to predict new entities based on the additional facts linked from the source (Section \ref{subsec:method_revision}).

\subsection{Generation with facts concatenation}
\label{subsec:method_concat}
Current summarization methods are dominated by sequence-to-sequence models such as BART and T5 which take a source document $x=(x_1,\ldots,x_n)$ as input and produce a generated summary $y=(y_1, \ldots, y_N)$ as output, where $y=f(x)$. Building on this paradigm, the most straightforward way to provide the model with additional external knowledge is to concatenate additional information to the source input being given to the model. In the case of our knowledge subgraph this yields $\hat{x} = concat([x, k_1, \dots, k_n])$ where $(k_1, \dots, k_n)$ is the linearization of the facts in the knowledge subgraph. 

In this case, we linearize facts into literal strings with "[SEP]" as the fact separator. For example, if the knowledge subgraph contains two facts: ("Simon Coveney", "country of citizen", "Ireland") and ("Taoiseach", "subclass of", "prime minister"), the linearized form of this knowledge subgraph is "[SEP] Simon Coveney country of citizen Ireland [SEP] Taoiseach subclass of prime minister [SEP]". As each linearized fact contains at least three tokens, the linearized facts for all entity categories drastically exceed the input length limitation of popular transformer-based summarization models.  We opt to train the model to correct only location type of entities, as they appear most frequently in the source (Figure~\ref{fig:entity_pie}) and have the best coverage improvement by fact linking (Figure~\ref{fig:entity_linking}).  Intuitively, the direct concatenation approach aims to teach the seq2seq model to learn to select useful facts in the source for the summary generation. 

Similar to the approach described in Section~\ref{sec:motivation}, we construct training data as follows. For each input document, we extract all entities in the source which are linkable to a Wikidata ID. We then construct the knowledge subgraph for this document by extracting all one-hop relations on the Wikidata knowledge graph that originate at any of the source entities. On average, this construction leads to 1837 facts in the knowledge subgraph for all source entities in the categories of \{Location, Person, Organization, event, Art, Consumer Good, Other\} per source document.

\begin{table}[t]
	\resizebox{0.49\textwidth}{!}{%
		\begin{tabular}{l|c|c|c}
			\toprule
			Input               & ROUGE-1        & ROUGE-L        & FactCC         \\ \midrule
			source only         & 43.67          & 36.25          & 23.71          \\ 
			\  + random words   & 44.40          & 36.50          & 23.72          \\ 
			\  + random  facts  & 44.59          & 36.54          & 24.11          \\ 
			\  + location facts & \textbf{44.83} & \textbf{36.76} & \textbf{24.15} \\\bottomrule
		\end{tabular}
	}
	\caption{Appending facts directly to the source using T5-3B model. }
	\label{tab:t5_appending}
\end{table}

Table ~\ref{tab:t5_appending} shows the results of appending location facts to the end of the source for XSUM.  
One interesting observation is that adding any information to the end of the source input can improve the ROUGE score as well as the FactCC scores.  Comparing the input of the original source or source appended with random words,\footnote{Obtained by sampling uniformly over the T5 vocabulary.} we can see that by simply concatenating random words, we were able to have higher ROUGE scores. Moreover, if we append random location facts or location facts that are linked from the source entities as additional input, we can see that the seq2seq model would be able to have higher ROUGE and FactCC scores, suggesting that these linked facts are useful for improving factual consistency of summaries when directly used in seq2seq models. 

\noindent\textbf{Limitations:}
Despite limiting the knowledge subgraph to contain only facts related to location entities, linearizing the facts and appending them to the end of the source often exceeds the length limit of the summarization models. Most of these models (e.g. BART and T5) have a length limit of 1024 \citep{lewis-etal-2020-bart,raffel2020exploring}. On average, we obtained 790 facts in the location-based knowledge subgraph. As each fact contains a triple with at least three tokens, this approach quickly becomes intractable. We set the T5 input token length to 1024 and facts that exceed this length will be automatically pruned. 

Using longer transformers such as Longformers \cite{beltagy2020longformer} or Big Bird \cite{zaheer2020big} can lessen, though not fully addressed, the length problem as such models usually have a default maximum length of 4096 tokens. In the knowledge subgraphs we obtained for XSUM, the average fact length is 7.8, Longformers can only accommodate around 500 facts, not counting the input text itself. As more source entities and larger numbers of facts are considered, this input context is quickly saturated. The facts could be further pruned using either heuristics or trained models, but this is itself a very difficult problem given the lack of direct supervision.

\subsection{Fact-based revision}
\label{subsec:method_revision}
As discussed in the previous section, direct concatenation of facts is not a scalable solution because the knowledge subgraph often exceeds the input length restriction of Transformers. In this section, we instead consider a two-stage generate-and-revise approach: \textbf{1)} first, a standard seq-to-seq model  
takes the source text as input and generates a target summary (Figure \ref{fig:overview1}-A). \textbf{2)} Next, an entity linking model 
identifies typed entities in the generated source. These entities are then masked out, producing a skeleton summary (Figure \ref{fig:overview1}-B). \textbf{3)} Finally, a fact-aware model 
is used to predict new entities to fill those masks resulting in a final summary (Figure \ref{fig:overview1}-D).\footnote{During training, we use the gold reference summary as the candidate summary and use system-generated summary during the inference. }

For steps 1 and 2, we use the same T5-3B and Google Cloud NLU models described at the beginning of Section \ref{sec:methods}. For step 3, we hypothesize that a model with the ability to access facts would produce more factual summaries that are consistent with external world knowledge. To confirm this hypothesis, we employ the Fact Injected Language Model (\fae{}) model \citep{verga-etal-2021-adaptable} for summarization revision. \fae{} is a neural language model that includes a “fact memory”, which stores an interpretable neuro-symbolic KB. Importantly, the model does not concatenate a seed set of facts to the input, but instead, stores them in a separate memory. The model learns to retrieve a small subset of facts from this memory and then incorporates those retrievals into its final prediction. This addresses the scaling issues of the previous section as the model can store millions of facts, learn to retrieve a set of relevant candidates, and incorporate that factual information into its predictions.

In brief, the input to the model is $x=concat([source, skeleton])$ which first produces hidden states $z=f(x)$. For hidden state $z_i$ corresponding to a mask appearing at the $i$th token, the model performs an attention over the fact memory as $a=g(z, M^{key})$. Each entry in $M^{key}$ corresponds to a single fact and is formed as a function of the subject and relation of that fact. The model then retrieves the corresponding values for the top K scoring fact keys, where each value is a function of the object set corresponding to the fact. For example a single fact would be $M^{key}_j = h([Barack\_Obama, born\_in])$ with the corresponding value being $M^{val}_j = \hat{h}(Hawaii)$. Finally, the model predicts an output entity as $\hat{y_i} = \hat{f}(z_i, a,  M^{val})$. Refer to \citet{verga-etal-2021-adaptable} for additional details on the \fae{} model.

\section{Results}
\label{sec:result}

In the following section, we discuss the results of using the revision model to specifically correct entity errors. In all the experiments, we add an entity correction baseline - T5 masking prediction (T5m) - that takes the concatenated source document and masked candidate summary and predicts the masks in a sequential fashion. We discuss the results with two parts: 1) oracle correction on gold-reference summaries 2) revision model based on system-generated summaries. 

We report the standard summarization evaluation metric - ROUGE \citep{lin-2004-rouge} and factual consistency evaluation metric - FactCC \citep{kryscinski-etal-2020-evaluating}. In addition, the entity correction models are evaluated based on two measures: entity correctness and entity consistency. Entity correctness matches predicted entities to the target entities, and  entity consistency matches predicted entities to the source entities. The entity matching is resolved by linking surface forms to their Wikidata IDs.

\subsection{Oracle correction}
To isolate the effects of the revision model away from the performance of the initial masked skeleton summary, we first evaluate the revision model on entity-masked gold target summaries. 
Table~\ref{tab:xsum_upper_bound_result} shows the oracle results of using the two different revision models. Note \cnndm abstractive subset is constructed by filtering document-summary pairs to contain the cases where targets contain entities that are not in the source by Wikidata IDs. We further divide all target entities into two categories: abstractive entities that are not in the source and extractive entities where the entity does appear in the source.

First, we can see that both revision models display different performances on abstractive and extractive entities. On XSUM, most of the abstractive entities can be corrected. For example, T5m can predict 73.1\% of abstractive entities correctly and \fae{} can predict 77.5\% of them correctly. However, this number drastically decreases on the \cnndm abstractive subset. In this setting, both models can only predict around 30\% of abstractive entities correctly, as the initial summarization model (i.e. T5-3B fine-tuned on \cnndm training set) learns an extractive strategy that prefers to predict source entities.  Nevertheless, if we compare only the abstractive entities on XSUM or \cnndm, we can see that \fae{}, which incorporates the additional knowledge subgraph, has a better performance on entity matching to the targets. On the other hand, T5m models perform better on extractive entities.

\begin{table}[t]
	\small
	\resizebox{0.48\textwidth}{!}{%
		\begin{tabular}{l|lll}
			\toprule
			Method & Abstractive    & Exttractive    & Full           \\ \toprule
			       &                & XSUM           &                \\ \midrule
			T5+T5m & 73.1           & \textbf{71.2}  & 72.4           \\ 
			\fae{} & \textbf{77.5}  & 67.2           & \textbf{74.5}  \\ \midrule
			       &                & \cnndm         &                \\ \midrule
			T5+T5m & 31.02          & \textbf{75.63} & \textbf{73.39} \\ 
			\fae{} & \textbf{33.75} & 73.24          & 68.44          \\ \midrule
		\end{tabular}
	}
	\caption{Upper bound results (entity ID matching) on XSUM and \cnndm abstractive subset: using gold-reference summary with MASK for entity prediction.}
	\label{tab:xsum_upper_bound_result}
\end{table}

\subsection{Revision model}

Table~\ref{tab:main_results_corectness} and Table~\ref{tab:main_results_consistency}  shows the results of using different revision models on the masked T5-generated candidate summary. In the revision settings, we divide the testing dataset into an abstractive subset and extractive subset. In this experiment, the abstractive subset contains document-summary pairs where the gold reference summary contains at least one entity that is not in the source. 

\paragraph{Abstractive metric:} In terms of entity correctness, we can see a similar trend to the oracle results.  Compared to using a T5 trained specifically on entities for error correction, using \fae{} with additional knowledge from entity linking results in higher correctness on abstractive entities. On XSUM, the entity matching with respect to the target increased to almost 5 percent with \fae{}, indicating a significant amount of external knowledge being used correctly. This result is expected, as XSUM contains many abstractive entities in the target (Figure~\ref{fig:entity_linking}). Similarly, on \cnndm, \fae{} is able to improve the entity matching in the abstractive subset by 0.8 percent, compared to the T5m correction model that has a decrease of 0.6 percent. This indicates that \fae{} is predicting more abstractive entities and T5m, on the contrary, is predicting more extractive entities. 

\begin{table}[t]
	\small
	\resizebox{0.49\textwidth}{!}{%
		\begin{tabular}{l|lll}
			\hline
			Method   & Abstractive    & Extractive     & Full           \\ 
			\toprule
			         &                & XSUM           &                \\
			\midrule
			T5       & 68.72          & 64.29          & 66.31          \\
			+ T5m    & 68.73          & 64.33          & 66.34          \\
			+ \fae{} & \textbf{73.40} & \textbf{65.32} & \textbf{71.60} \\
			\midrule
			         &                & \cnndm         &                \\
			\midrule
			T5       & 29.58          & 72.45          & 66.85          \\
			+ T5m    & 28.95          & \textbf{74.88} & \textbf{67.15} \\
			+ \fae{} & \textbf{30.31} & 72.25          & 66.71          \\ \bottomrule
		\end{tabular}
	}
	\caption{Results of using \fae{} for error correction on T5 outputs on XSUM. We report correctness by measuring the entity ID matching between targets and model predictions.}
	\label{tab:main_results_corectness}
\end{table}

In terms of overall performance on XSUM, when compared to T5m, our model also has higher correctness on extractive entities and achieves overall higher correctness. However, on \cnndm, since the majority of entities are extractive (72.8 \%) compared to (33.5\%) on XSUM, and T5m is performing better on extractive entities, the overall correctness of T5m is better on \cnndm.

\paragraph{Extractive metric:} In Table~\ref{tab:main_results_consistency}, we present the results of factual consistency measures - entity consistency (Consistency) and FactCC.  When looking at the first column by matching the generated entities to the source, we can see that compared to the original T5 outputs, T5m produces more extractive entities and our model seems to produce fewer entities that match the source. \fae{} seems to make the generated summary more abstractive, with more words inferred from the source, but not directly appearing in the source. 

It is important to note again that these extractive metrics explicitly penalize predicting strings that do not appear in the source \textit{even if that string is correct}, as in the case of an out of article entity. 

\begin{table}[t]
	\small
	\resizebox{0.5\textwidth}{!}{%
		
		\begin{tabular}{l|lll}
			\hline
			Method   & Consistency    & FactCC         & ROUGE-1        \\
			\toprule
			         &                & XSUM           &                \\
			\midrule
			T5       & 73.85          & \textbf{22.84} & 45.14          \\
			+ T5m    & \textbf{74.21} & 21.32          & \textbf{45.21} \\
			+ \fae{} & 73.15          & 21.32          & 45.09          \\
			\midrule
			         &                & \cnndm         &                \\
			\midrule
			T5       & 84.12          & 69.22          & 44.32          \\
			+ T5m    & \textbf{85.31} & \textbf{71.02} & \textbf{44.67} \\
			+ \fae{} & 83.57          & 68.51          & 43.81          \\ \bottomrule
		\end{tabular}
	}
	\caption{Results for T5 with and without revision models on XSUM. Consistency and FactCC both measure extractiveness by comparing to the source. We also report ROUGE-1; however, the ROUGE-1 of \fae{} is calculated by matching the canonical form of entities to the gold-reference target. This can often result in a string mismatch with the target summary (variance between an entity's surface form and its canonical name) and the model being penalized despite the underlying entities being the same. }
	\label{tab:main_results_consistency}
\end{table}

\subsubsection{Human evaluation}
We also conducted a small human evaluation to check whether error correction with the external facts would improve the  summary in terms of faithfulness. We randomly sampled 100 system-generated summaries that contain locations and use \fae{} to perform entity revisions. We present three annotators with the masked system-generated summary, entities before and after the correction in randomized order, and ask them to choose from the following four options based on the faithfulness to the target\footnote{We encourage the annotators to check on search engines if an entity is not directly supported by the source or the target.}: 1) entity A is better 2) entity B is better 3) both are good 4) both are bad.  

Table~\ref{tab:human_eval} presents the human evaluation results. On average, we see a 4.4\% of faithfulness improvement after the revision based on human judges.

\begin{table}[h]
	\centering
	\resizebox{0.5\textwidth}{!}{%
		\begin{tabular}{c|c|c|c|c}
			            & Revised & Original & Both Good & Both Bad \\ \midrule
			avg. rating & 26.7\%  & 22.3\%   & 17.7\%    & 31.3\%   \\
			       
		\end{tabular}
	}
	\caption{Average rating based on 96 samples\footnotemark~  before and after revision by three different annotators. The annotators agreements are: Kappa 0.7392,
		Fleiss 0.7391,
		Alpha 0.7397,
	Scotts 0.7388.}
	\label{tab:human_eval}
\end{table}

\footnotetext{Four examples are dropped due to entity extraction errors.~}
\section{Analysis and discussion}

\paragraph{Do knowledge aware models perform differently by category?}

\begin{table}[t]
	\resizebox{0.5\textwidth}{!}{%
		\begin{tabular}[\columnwidth]{l|ll|ll}
			\toprule
			Dev.         & Extractive & Abstractive & In Source & Only in KB \\ \midrule
			location     & 0.95       & 0.70        & 40\%      & 19\%       \\ 
			person       & 0.75       & 0.65        & 29\%      & 8\%        \\ 
			organization & 0.50       & 0.40        & 40\%      & 13\%       \\
			event        & 0.62       & 0.35        & 21\%      & 12\%       \\ 
			art          & 0.59       & 0.14        & 32\%      & 15\%       \\
			other        & 0.25       & 0.30        & 34\%      & 7\%        \\ 
			\bottomrule
		\end{tabular}
	}
	\caption{Results of \fae{} correction accuracy by categories.}
	\label{tab:result_categories}
\end{table}

While our primary experiments focused on location entities, we also analyzed the performance of several other entity categories using \fae{} as our revision model. We divide target entities into two subsets: abstractive entities refer to the target entities that do not appear in the source, and extractive entities can be found in the source. 

The right-most column of Table~\ref{tab:result_categories} shows the percentage of abstractive target entities (entities that are in the target but not in the source) that can be found in the knowledge subgraph linked from the source entities or in the source. When compared to different categories, frequent entity types such as location, person, and organization have a better prediction accuracy on both extractive and abstractive entities. However, the accuracy of entities, other than location is relatively low. This is possibly due to less coverage in the knowledge base or greater difficulty in identifying relevant facts. The low oracle entity prediction accuracy results also translate into the pipeline experiments: if we perform entity correction based on person or organization, \fae{} actually reduces the faithfulness of the generated summary, indicating the importance of having well-aligned external knowledge. 

\begin{table}[h]
	\resizebox{0.5\textwidth}{!}{%
		\begin{tabular}{llll}
			\toprule
			Before/After & XSUM & Correct               & Wrong        \\ \midrule
			
			person       & abs  & \textbf{79.95}/76.07  & 20.05/ 23.93 \\
			             & ext  & \textbf{79.80}/ 72.63 & 20.20/27.36  \\
			organization & abs  & \textbf{66.80}/ 50.08 & 33.20/49.92  \\
			             & ext  & \textbf{73.96}/55.50  & 26.04/44.50  \\  
			\bottomrule
		\end{tabular}
	}%
	\caption{Results of using \fae{} for error correction on t5 outputs on XSUM (person and organization categories).} 
\end{table}

\section{Related work}
\paragraph{Factuality in summarization}
Factual consistency of summarization has drawn much research interest since the proposal of FactCC \cite{kryscinski-etal-2020-evaluating}, an evaluation model that classifies the generated summary as factually consistent/inconsistent to the source. 
Later, several question answering-based summarization evaluation methods were proposed \cite{wang-etal-2020-asking, durmus-etal-2020-feqa,nan-etal-2021-entity,zeng-etal-2021-gradient}. These models measure the factual consistency by matching the entity answers that are produced by a QA model with inputs of (question, source) and (question, generated summary). 

Recently, many methods have also been proposed to improve the factual consistency of generated summaries. The majority of these models reduces the probability of generating novel entities by imposing constraint w.r.t the source, such as quantity entity matching \citep{zhao-etal-2020-reducing}, intermediate planning with entity chains \citep{narayan2021planning}, or simple filtering
\citep{nan-etal-2021-entity}.  \citet{filippova-2020-controlled} controls hallucinations with unconditional and conditional LMs. \citet{dong-etal-2020-multi,cao-etal-2020-factual} propose to increase factual consistency by post-error corrections with QA-based models or BART. \citet{cao2018faithful,zhu-etal-2021-enhancing} utilize dependency parsing tools to identify and match the relations in an input document and its summary.

\paragraph{Generation with factual information}
Many recent models have been proposed for retrieval augmented language models using passages \citep{guu2020realm, lewis2020retrieval}, mentions \citep{sun2021reasoning}, and facts \citep{verga-etal-2021-adaptable}. In this paper, we experiment with incorporating facts that are directly linked to the entities in the source. Several models have been proposed to combine symbolically interpretable factual information and subsymbolic neural knowledge \citep{Cohen2020Scalable, Ren2020Query2box,DBLP:journals/corr/abs-2104-07606}.

\section{Conclusion}
In this paper, we have discovered that a large portion of so-called external hallucinations in text summarization can be verified by external knowledge, obtained by linking entities from the Freebase. We have shown multiple ways to explore how to combine this knowledge into a faithful generation of summaries. Our work not only proposes a pipeline that can guarantee better faithfulness but also discuss some valuable insights about current limitations and promising directions for knowledge-grounded text generation.

\bibliography{anthology,custom}

\begin{thebibliography}{37}
\expandafter\ifx\csname natexlab\endcsname\relax\def\natexlab#1{#1}\fi

\bibitem[{Beltagy et~al.(2020)Beltagy, Peters, and
  Cohan}]{beltagy2020longformer}
Iz~Beltagy, Matthew~E Peters, and Arman Cohan. 2020.
\newblock Longformer: The long-document transformer.
\newblock \emph{arXiv preprint arXiv:2004.05150}.

\bibitem[{Cao et~al.(2020)Cao, Dong, Wu, and Cheung}]{cao-etal-2020-factual}
Meng Cao, Yue Dong, Jiapeng Wu, and Jackie Chi~Kit Cheung. 2020.
\newblock \href {https://doi.org/10.18653/v1/2020.emnlp-main.506} {Factual
  error correction for abstractive summarization models}.
\newblock In \emph{Proceedings of the 2020 Conference on Empirical Methods in
  Natural Language Processing (EMNLP)}, pages 6251--6258, Online. Association
  for Computational Linguistics.

\bibitem[{Cao et~al.(2018)Cao, Wei, Li, and Li}]{cao2018faithful}
Ziqiang Cao, Furu Wei, Wenjie Li, and Sujian Li. 2018.
\newblock Faithful to the original: Fact aware neural abstractive
  summarization.
\newblock In \emph{Proceedings of the AAAI Conference on Artificial
  Intelligence}.

\bibitem[{Cohen et~al.(2020)Cohen, Sun, Hofer, and Siegler}]{Cohen2020Scalable}
William~W. Cohen, Haitian Sun, R.~Alex Hofer, and Matthew Siegler. 2020.
\newblock \href {https://openreview.net/forum?id=BJlguT4YPr} {Scalable neural
  methods for reasoning with a symbolic knowledge base}.
\newblock In \emph{International Conference on Learning Representations}.

\bibitem[{Dong et~al.(2020)Dong, Wang, Gan, Cheng, Cheung, and
  Liu}]{dong-etal-2020-multi}
Yue Dong, Shuohang Wang, Zhe Gan, Yu~Cheng, Jackie Chi~Kit Cheung, and Jingjing
  Liu. 2020.
\newblock \href {https://doi.org/10.18653/v1/2020.emnlp-main.749} {Multi-fact
  correction in abstractive text summarization}.
\newblock In \emph{Proceedings of the 2020 Conference on Empirical Methods in
  Natural Language Processing (EMNLP)}, pages 9320--9331, Online. Association
  for Computational Linguistics.

\bibitem[{Durmus et~al.(2020)Durmus, He, and Diab}]{durmus-etal-2020-feqa}
Esin Durmus, He~He, and Mona Diab. 2020.
\newblock \href {https://doi.org/10.18653/v1/2020.acl-main.454} {{FEQA}: A
  question answering evaluation framework for faithfulness assessment in
  abstractive summarization}.
\newblock In \emph{Proceedings of the 58th Annual Meeting of the Association
  for Computational Linguistics}, pages 5055--5070, Online. Association for
  Computational Linguistics.

\bibitem[{Ebisu and Ichise(2019)}]{ebisu-ichise-2019-graph}
Takuma Ebisu and Ryutaro Ichise. 2019.
\newblock \href {https://doi.org/10.18653/v1/N19-1104} {Graph pattern entity
  ranking model for knowledge graph completion}.
\newblock In \emph{Proceedings of the 2019 Conference of the North {A}merican
  Chapter of the Association for Computational Linguistics: Human Language
  Technologies, Volume 1 (Long and Short Papers)}, pages 988--997, Minneapolis,
  Minnesota. Association for Computational Linguistics.

\bibitem[{Falke et~al.(2019)Falke, Ribeiro, Utama, Dagan, and
  Gurevych}]{falke-etal-2019-ranking}
Tobias Falke, Leonardo F.~R. Ribeiro, Prasetya~Ajie Utama, Ido Dagan, and Iryna
  Gurevych. 2019.
\newblock \href {https://doi.org/10.18653/v1/P19-1213} {Ranking generated
  summaries by correctness: An interesting but challenging application for
  natural language inference}.
\newblock In \emph{Proceedings of the 57th Annual Meeting of the Association
  for Computational Linguistics}, pages 2214--2220, Florence, Italy.
  Association for Computational Linguistics.

\bibitem[{Filippova(2020)}]{filippova-2020-controlled}
Katja Filippova. 2020.
\newblock \href {https://doi.org/10.18653/v1/2020.findings-emnlp.76}
  {Controlled hallucinations: Learning to generate faithfully from noisy data}.
\newblock In \emph{Findings of the Association for Computational Linguistics:
  EMNLP 2020}, pages 864--870, Online. Association for Computational
  Linguistics.

\bibitem[{Guu et~al.(2020)Guu, Lee, Tung, Pasupat, and Chang}]{guu2020realm}
Kelvin Guu, Kenton Lee, Zora Tung, Panupong Pasupat, and Ming-Wei Chang. 2020.
\newblock Realm: Retrieval-augmented language model pre-training.
\newblock \emph{arXiv preprint arXiv:2002.08909}.

\bibitem[{Hermann et~al.(2015)Hermann, Kocisky, Grefenstette, Espeholt, Kay,
  Suleyman, and Blunsom}]{Hermann2015cnndm}
Karl~Moritz Hermann, Tomas Kocisky, Edward Grefenstette, Lasse Espeholt, Will
  Kay, Mustafa Suleyman, and Phil Blunsom. 2015.
\newblock Teaching machines to read and comprehend.
\newblock In \emph{NIPS}.

\bibitem[{Kryscinski et~al.(2020)Kryscinski, McCann, Xiong, and
  Socher}]{kryscinski-etal-2020-evaluating}
Wojciech Kryscinski, Bryan McCann, Caiming Xiong, and Richard Socher. 2020.
\newblock \href {https://doi.org/10.18653/v1/2020.emnlp-main.750} {Evaluating
  the factual consistency of abstractive text summarization}.
\newblock In \emph{Proceedings of the 2020 Conference on Empirical Methods in
  Natural Language Processing (EMNLP)}, pages 9332--9346, Online. Association
  for Computational Linguistics.

\bibitem[{Ladhak et~al.(2021)Ladhak, Durmus, He, Cardie, and
  McKeown}]{ladhak2021faithful}
Faisal Ladhak, Esin Durmus, He~He, Claire Cardie, and Kathleen McKeown. 2021.
\newblock Faithful or extractive? on mitigating the
  faithfulness-abstractiveness trade-off in abstractive summarization.
\newblock \emph{arXiv preprint arXiv:2108.13684}.

\bibitem[{Lewis et~al.(2020{\natexlab{a}})Lewis, Liu, Goyal, Ghazvininejad,
  Mohamed, Levy, Stoyanov, and Zettlemoyer}]{lewis-etal-2020-bart}
Mike Lewis, Yinhan Liu, Naman Goyal, Marjan Ghazvininejad, Abdelrahman Mohamed,
  Omer Levy, Veselin Stoyanov, and Luke Zettlemoyer. 2020{\natexlab{a}}.
\newblock \href {https://doi.org/10.18653/v1/2020.acl-main.703} {{BART}:
  Denoising sequence-to-sequence pre-training for natural language generation,
  translation, and comprehension}.
\newblock In \emph{Proceedings of the 58th Annual Meeting of the Association
  for Computational Linguistics}, pages 7871--7880, Online. Association for
  Computational Linguistics.

\bibitem[{Lewis et~al.(2020{\natexlab{b}})Lewis, Perez, Piktus, Petroni,
  Karpukhin, Goyal, K{\"u}ttler, Lewis, Yih, Rockt{\"a}schel
  et~al.}]{lewis2020retrieval}
Patrick Lewis, Ethan Perez, Aleksandra Piktus, Fabio Petroni, Vladimir
  Karpukhin, Naman Goyal, Heinrich K{\"u}ttler, Mike Lewis, Wen-tau Yih, Tim
  Rockt{\"a}schel, et~al. 2020{\natexlab{b}}.
\newblock Retrieval-augmented generation for knowledge-intensive nlp tasks.
\newblock \emph{arXiv preprint arXiv:2005.11401}.

\bibitem[{Lin(2004)}]{lin-2004-rouge}
Chin-Yew Lin. 2004.
\newblock \href {https://aclanthology.org/W04-1013} {{ROUGE}: A package for
  automatic evaluation of summaries}.
\newblock In \emph{Text Summarization Branches Out}, pages 74--81, Barcelona,
  Spain. Association for Computational Linguistics.

\bibitem[{Lin et~al.(2018)Lin, Socher, and Xiong}]{lin-etal-2018-multi}
Xi~Victoria Lin, Richard Socher, and Caiming Xiong. 2018.
\newblock \href {https://doi.org/10.18653/v1/D18-1362} {Multi-hop knowledge
  graph reasoning with reward shaping}.
\newblock In \emph{Proceedings of the 2018 Conference on Empirical Methods in
  Natural Language Processing}, pages 3243--3253, Brussels, Belgium.
  Association for Computational Linguistics.

\bibitem[{Mao et~al.(2020)Mao, Ren, Ji, and Han}]{mao2020constrained}
Yuning Mao, Xiang Ren, Heng Ji, and Jiawei Han. 2020.
\newblock Constrained abstractive summarization: Preserving factual consistency
  with constrained generation.
\newblock \emph{arXiv preprint arXiv:2010.12723}.

\bibitem[{Maynez et~al.(2020)Maynez, Narayan, Bohnet, and
  McDonald}]{maynez-etal-2020-faithfulness}
Joshua Maynez, Shashi Narayan, Bernd Bohnet, and Ryan McDonald. 2020.
\newblock \href {https://doi.org/10.18653/v1/2020.acl-main.173} {On
  faithfulness and factuality in abstractive summarization}.
\newblock In \emph{Proceedings of the 58th Annual Meeting of the Association
  for Computational Linguistics}, pages 1906--1919, Online. Association for
  Computational Linguistics.

\bibitem[{Min et~al.(2013)Min, Grishman, Wan, Wang, and
  Gondek}]{min-etal-2013-distant}
Bonan Min, Ralph Grishman, Li~Wan, Chang Wang, and David Gondek. 2013.
\newblock \href {https://aclanthology.org/N13-1095} {Distant supervision for
  relation extraction with an incomplete knowledge base}.
\newblock In \emph{Proceedings of the 2013 Conference of the North {A}merican
  Chapter of the Association for Computational Linguistics: Human Language
  Technologies}, pages 777--782, Atlanta, Georgia. Association for
  Computational Linguistics.

\bibitem[{Nallapati et~al.(2016)Nallapati, Zhou, dos Santos, Gu̇l{\c{c}}ehre,
  and Xiang}]{nallapati-etal-2016-abstractive}
Ramesh Nallapati, Bowen Zhou, Cicero dos Santos, {\c{C}}a{\u{g}}lar
  Gu̇l{\c{c}}ehre, and Bing Xiang. 2016.
\newblock \href {https://doi.org/10.18653/v1/K16-1028} {Abstractive text
  summarization using sequence-to-sequence {RNN}s and beyond}.
\newblock In \emph{Proceedings of The 20th {SIGNLL} Conference on Computational
  Natural Language Learning}, pages 280--290, Berlin, Germany. Association for
  Computational Linguistics.

\bibitem[{Nan et~al.(2021)Nan, Nallapati, Wang, Nogueira~dos Santos, Zhu,
  Zhang, McKeown, and Xiang}]{nan-etal-2021-entity}
Feng Nan, Ramesh Nallapati, Zhiguo Wang, Cicero Nogueira~dos Santos, Henghui
  Zhu, Dejiao Zhang, Kathleen McKeown, and Bing Xiang. 2021.
\newblock \href {https://doi.org/10.18653/v1/2021.eacl-main.235} {Entity-level
  factual consistency of abstractive text summarization}.
\newblock In \emph{Proceedings of the 16th Conference of the European Chapter
  of the Association for Computational Linguistics: Main Volume}, pages
  2727--2733, Online. Association for Computational Linguistics.

\bibitem[{Narayan et~al.(2018)Narayan, Cohen, and
  Lapata}]{narayan-etal-2018-dont}
Shashi Narayan, Shay~B. Cohen, and Mirella Lapata. 2018.
\newblock \href {https://doi.org/10.18653/v1/D18-1206} {Don{'}t give me the
  details, just the summary! topic-aware convolutional neural networks for
  extreme summarization}.
\newblock In \emph{Proceedings of the 2018 Conference on Empirical Methods in
  Natural Language Processing}, pages 1797--1807, Brussels, Belgium.
  Association for Computational Linguistics.

\bibitem[{Narayan et~al.(2021{\natexlab{a}})Narayan, Zhao, Maynez,
  Sim{\~{o}}es, and McDonald}]{DBLP:journals/corr/abs-2104-07606}
Shashi Narayan, Yao Zhao, Joshua Maynez, Gon{\c{c}}alo Sim{\~{o}}es, and
  Ryan~T. McDonald. 2021{\natexlab{a}}.
\newblock \href {http://arxiv.org/abs/2104.07606} {Planning with entity chains
  for abstractive summarization}.
\newblock \emph{CoRR}, abs/2104.07606.

\bibitem[{Narayan et~al.(2021{\natexlab{b}})Narayan, Zhao, Maynez, Sim{\~o}es,
  Nikolaev, and McDonald}]{narayan2021planning}
Shashi Narayan, Yao Zhao, Joshua Maynez, Gon{\c{c}}alo Sim{\~o}es, Vitaly
  Nikolaev, and Ryan McDonald. 2021{\natexlab{b}}.
\newblock Planning with learned entity prompts for abstractive summarization.
\newblock \emph{Transactions of the Association for Computational Linguistics},
  9:1475--1492.

\bibitem[{Pagnoni et~al.(2021)Pagnoni, Balachandran, and
  Tsvetkov}]{pagnoni-etal-2021-understanding}
Artidoro Pagnoni, Vidhisha Balachandran, and Yulia Tsvetkov. 2021.
\newblock \href {https://doi.org/10.18653/v1/2021.naacl-main.383}
  {Understanding factuality in abstractive summarization with {FRANK}: A
  benchmark for factuality metrics}.
\newblock In \emph{Proceedings of the 2021 Conference of the North American
  Chapter of the Association for Computational Linguistics: Human Language
  Technologies}, pages 4812--4829, Online. Association for Computational
  Linguistics.

\bibitem[{Raffel et~al.(2020)Raffel, Shazeer, Roberts, Lee, Narang, Matena,
  Zhou, Li, and Liu}]{raffel2020exploring}
Colin Raffel, Noam Shazeer, Adam Roberts, Katherine Lee, Sharan Narang, Michael
  Matena, Yanqi Zhou, Wei Li, and Peter~J Liu. 2020.
\newblock Exploring the limits of transfer learning with a unified text-to-text
  transformer.
\newblock \emph{Journal of Machine Learning Research}, 21:1--67.

\bibitem[{Ren et~al.(2020)Ren, Hu, and Leskovec}]{Ren2020Query2box}
Hongyu Ren, Weihua Hu, and Jure Leskovec. 2020.
\newblock \href {https://openreview.net/forum?id=BJgr4kSFDS} {Query2box:
  Reasoning over knowledge graphs in vector space using box embeddings}.
\newblock In \emph{International Conference on Learning Representations}.

\bibitem[{See et~al.(2017)See, Liu, and Manning}]{see-etal-2017-get}
Abigail See, Peter~J. Liu, and Christopher~D. Manning. 2017.
\newblock \href {https://doi.org/10.18653/v1/P17-1099} {Get to the point:
  Summarization with pointer-generator networks}.
\newblock In \emph{Proceedings of the 55th Annual Meeting of the Association
  for Computational Linguistics (Volume 1: Long Papers)}, pages 1073--1083,
  Vancouver, Canada. Association for Computational Linguistics.

\bibitem[{Sun et~al.(2021)Sun, Verga, Dhingra, Salakhutdinov, and
  Cohen}]{sun2021reasoning}
Haitian Sun, Pat Verga, Bhuwan Dhingra, Ruslan Salakhutdinov, and William~W
  Cohen. 2021.
\newblock Reasoning over virtual knowledge bases with open predicate relations.
\newblock \emph{International Conference on Machine Learning}.

\bibitem[{Verga et~al.(2021)Verga, Sun, Baldini~Soares, and
  Cohen}]{verga-etal-2021-adaptable}
Pat Verga, Haitian Sun, Livio Baldini~Soares, and William Cohen. 2021.
\newblock \href {https://doi.org/10.18653/v1/2021.naacl-main.288} {Adaptable
  and interpretable neural {M}emory{O}ver symbolic knowledge}.
\newblock In \emph{Proceedings of the 2021 Conference of the North American
  Chapter of the Association for Computational Linguistics: Human Language
  Technologies}, pages 3678--3691, Online. Association for Computational
  Linguistics.

\bibitem[{Vrande{\v{c}}i{\'c} and Kr{\"o}tzsch(2014)}]{vrandevcic2014wikidata}
Denny Vrande{\v{c}}i{\'c} and Markus Kr{\"o}tzsch. 2014.
\newblock Wikidata: a free collaborative knowledgebase.
\newblock \emph{Communications of the ACM}, 57(10):78--85.

\bibitem[{Wang et~al.(2020)Wang, Cho, and Lewis}]{wang-etal-2020-asking}
Alex Wang, Kyunghyun Cho, and Mike Lewis. 2020.
\newblock \href {https://doi.org/10.18653/v1/2020.acl-main.450} {Asking and
  answering questions to evaluate the factual consistency of summaries}.
\newblock In \emph{Proceedings of the 58th Annual Meeting of the Association
  for Computational Linguistics}, pages 5008--5020, Online. Association for
  Computational Linguistics.

\bibitem[{Zaheer et~al.(2020)Zaheer, Guruganesh, Dubey, Ainslie, Alberti,
  Ontanon, Pham, Ravula, Wang, Yang et~al.}]{zaheer2020big}
Manzil Zaheer, Guru Guruganesh, Kumar~Avinava Dubey, Joshua Ainslie, Chris
  Alberti, Santiago Ontanon, Philip Pham, Anirudh Ravula, Qifan Wang, Li~Yang,
  et~al. 2020.
\newblock Big bird: Transformers for longer sequences.
\newblock In \emph{NeurIPS}.

\bibitem[{Zeng et~al.(2021)Zeng, Chen, Xu, and Li}]{zeng-etal-2021-gradient}
Zhiyuan Zeng, Jiaze Chen, Weiran Xu, and Lei Li. 2021.
\newblock \href {https://aclanthology.org/2021.emnlp-main.337} {Gradient-based
  adversarial factual consistency evaluation for abstractive summarization}.
\newblock In \emph{Proceedings of the 2021 Conference on Empirical Methods in
  Natural Language Processing}, pages 4102--4108, Online and Punta Cana,
  Dominican Republic. Association for Computational Linguistics.

\bibitem[{Zhao et~al.(2020)Zhao, Cohen, and Webber}]{zhao-etal-2020-reducing}
Zheng Zhao, Shay~B. Cohen, and Bonnie Webber. 2020.
\newblock \href {https://doi.org/10.18653/v1/2020.findings-emnlp.203} {Reducing
  quantity hallucinations in abstractive summarization}.
\newblock In \emph{Findings of the Association for Computational Linguistics:
  EMNLP 2020}, pages 2237--2249, Online. Association for Computational
  Linguistics.

\bibitem[{Zhu et~al.(2021)Zhu, Hinthorn, Xu, Zeng, Zeng, Huang, and
  Jiang}]{zhu-etal-2021-enhancing}
Chenguang Zhu, William Hinthorn, Ruochen Xu, Qingkai Zeng, Michael Zeng,
  Xuedong Huang, and Meng Jiang. 2021.
\newblock \href {https://doi.org/10.18653/v1/2021.naacl-main.58} {Enhancing
  factual consistency of abstractive summarization}.
\newblock In \emph{Proceedings of the 2021 Conference of the North American
  Chapter of the Association for Computational Linguistics: Human Language
  Technologies}, pages 718--733, Online. Association for Computational
  Linguistics.

\end{thebibliography}
\bibliographystyle{acl_natbib}

\newpage
\appendix
\section{Appendix}
\label{sec:appendix}
\subsection{Datasets}
Table \ref{tab:data} shows the statistics of the datasets used in our experiments. Note that summarization models finetuned on CNNDM often favor an extractive strategy \citep{see-etal-2017-get}. To encourage the fine-tuned summarization model to produce more out-of-article entities, we filter an abstractive subset of CNN/Daily Mail dataset \citep{Hermann2015cnndm,nallapati-etal-2016-abstractive}, noted as CNNDM$_{abs}$, where at least one location entity in the target is not in the source. We obtained 95387/4357/3769 data instances for train/dev/test on \cnndm. 

\begin{table}[h]
	\small
	\resizebox{0.48\textwidth}{!}{%
		\begin{tabular}{l|lll}
			\toprule
			Dataset & Train   & Dev    & Test   \\ \toprule
			XSUM    & 204,045 & 11,332 & 11,334 \\ 
			CNNDM   & 287,226 & 13,368 & 11,490 \\ 
			\cnndm  & 95,387  & 4,357  & 3,769  \\
			\bottomrule
		\end{tabular}
	}
	\caption{Dataset statistics in terms of number of examples in train, dev, and test splits for three summarization datasets used in our experiments.}
	\label{tab:data}
\end{table}

\subsection{Finetuning \fae{} for summarization tasks}
We modify the entity correction task in summarization as an open-domain question answering task, which \fae{} is designed for. The setup is as follows. We treat the source document as the context and the masked skeleton sentence, obtained from masking entities in either the gold reference summary or system-generated summary, as the question. \fae{} learns to extract useful information from the open domain (knowledge base) to provide evidence for the entity prediction. We focus on a subset of entities that are answerable using entities from the knowledge base. For example, the answer “United States” is an entity in Wikidata whose identity is Q30.

Same as described in \citet{verga-etal-2021-adaptable}, at finetuning time, we freeze entity embeddings $\mathbb{E}$ and relation embeddings $\mathbb{R}$. All transformer layers with four transformation matrices are finetuned with the loss: 

$$\text{loss}_{\text{finetune}} = \text{loss}_{\text{fact}} + \text{loss}_{\text{ans}}.$$

The number of base parameters, including the encoder and decoder transformer
parameters and the finetuning optimizer, is derived from the original papers. We set the max length of \fae{} to 512, as it only needs the skeleton summary and the original source document as the input for entity correction. The \fae{} models for XSUM and \cnndm are trained on Google Cloud TPU v3 Pod with 128-core Pod slices.

\subsection{Example of usefulness of abstractive entity}
\begin{table}[h]
	\small
	\begin{tabular}{p{1cm}|p{6cm}}
		\toprule
		Source & \indent {\color[HTML]{008000} Mr. Cowen} had to deny being drunk or hungover during the RTE interview. The {\color{red} taoiseach} was interviewed live from his party's conference, which is taking place in Galway. … I would hate to think the reputation of the country or the office of {\color{red} taoiseach} would in any way be affected by what I had to say." {\color[HTML]{008000} Mr. Cowen} again denied any suggestions he was hungover. ... Simon Coveney, also of Fine Gael, who said in a Twitter message on Tuesday that {\color[HTML]{008000} Mr. Cowen} sounded "half-way between drunk and hungover" in the interview, has said he accepted the {\color{red} taoiseach}'s apology. … \\ \midrule
		Target & \indent {\color[HTML]{FF7E00} Irish Prime Minister} {\color[HTML]{007FFF}Brian Cowen} has admitted that a controversial radio interview he gave on Tuesday was not his "best performance". \\ \midrule
		Gen.   & \indent {\color{red} Taoiseach }{\color[HTML]{FF7E00} Irish Prime Minister} {\color[HTML]{007FFF}Brian Cowen} has apologised for the " hoarseness" of his voice in an interview on Tuesday.\\                           \bottomrule                                                         
	\end{tabular} 
	\caption{The target summary contains out-of-article entities – {\color[HTML]{FF7E00} Irish Prime Minister} and {\color[HTML]{007FFF}Brian Cowen} – that are important to be included in the summary. We can see that a summarization model is able to generate this example successfully with additional world knowledge that “{\color{red}Taoiseach}” is equivalent to “{\color[HTML]{FF7E00} Irish Prime Minister}” and “{\color{red}Taoiseach} {\color[HTML]{008000} Mr. Cowen}” refers to “{\color[HTML]{007FFF}Brian Cowen}”.}
	\label{tab:open_example1}
\end{table}

\begin{figure}[h]
	\centering
	\includegraphics[width=.99\linewidth]{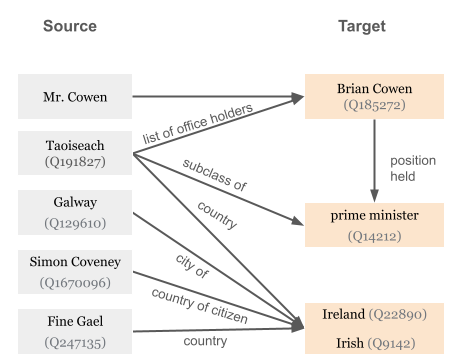}
	\caption{Example of facts in Wikidata KB that connect the source entities to abstractive target entities (entities that do not appear in the source). }
	\label{fig:kb}
\end{figure}

\begin{figure*}[h]
	\centering
	\includegraphics[width=.99\linewidth]{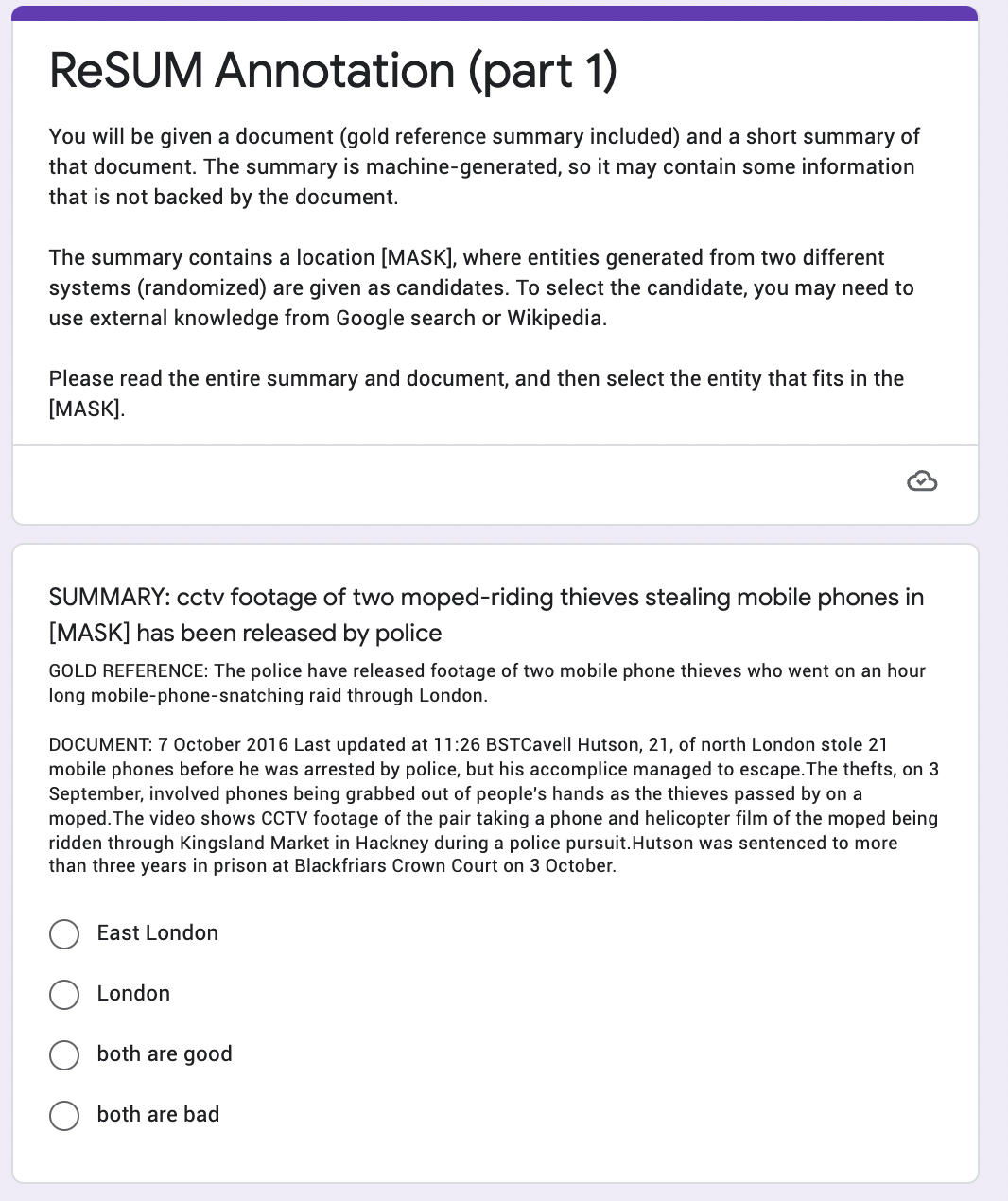}
	\caption{An example of human annotation template. }
	\label{fig:annotation}
\end{figure*}

\end{document}